\documentclass[runningheads]{llncs}
\usepackage[T1]{fontenc}
\usepackage{graphicx,verbatim}
\usepackage{amsmath,amssymb}
\usepackage{booktabs}
\usepackage{url}
\usepackage{multirow}

\usepackage{color}
\usepackage{lineno}
\modulolinenumbers[1]
\begin{document}
%

\title{On-Device Multi-Species Malaria Detection with Uncertainty-Calibrated Slide-Level Aggregation}
\titlerunning{On-Device Multi-Species Malaria Detection}

\author{Idaya Seidu\inst{1} \and
Ahmed Tahiru Issah\inst{1} \and
Charles B. Delahunt\inst{2} \and
Carine Mukamakuza\inst{1}\thanks{Corresponding author}}

\authorrunning{I. Seidu et al.}

\institute{Carnegie Mellon University Africa, Kigali Innovation City, Kigali, Rwanda \\
\email{\{iseidu,aissah,cmukamak\}@andrew.cmu.edu}
\and
University of Washington, Seattle, Washington, USA \\
\email{delahunt@uw.edu}
}

\maketitle
\begin{abstract}
Malaria remains a leading cause of mortality in resource-limited settings, where expert microscopists are scarce.
Automated diagnosis based on microscopy images thus has strong potential to improve care delivery. But for an algorithm to deploy, a necessary requirement is that it meet a suite of non-obvious (from a machine learning (ML) perspective) clinical constraints. Therefore, in close consultation with a national health center we developed a malaria diagnosis pipeline  which addresses key requirements listed by the health care center but typically ignored in the ML malaria literature. In particular, it includes: (i) stopping criteria (to reduce image acquisition and time-to-result); (ii) human-in-the-loop functionality (for review and accountability); (iii) multi-species discrimination (since treatment varies by species); (iv) thick film detection (standard for microscopy); (v) computationally-efficient uncertainty calculations (to aid clinician review); and (vi) an edge device platform (since internet can be spotty in this catchment area).
The mobile system
performs all inference on-device using YOLOv13n deployed via TensorFlow Lite.
It detects four species 
and white blood cells from Giemsa-stained thick blood smear images, aggregating per-image detections into slide-level parasitemia with World Health Organization (WHO)-standard quantification.
This paper highlights these various clinical constraints and offers methods to address them.
Evaluated on 2,739 annotated images across all four species, the system achieves mAP@0.5 of 0.863, per-image parasite count correlation of $r = 0.812$, slide-level $r = 0.951$ (soft counting, 10 images/slide), and runs entirely offline with a pipeline time of $10.27 \pm 1.65$~s per image.

\end{abstract}
\section{Introduction}
\label{sec:intro}

Malaria is responsible for over 600,000 deaths annually, with 95\% of cases in sub-Saharan Africa~\cite{who2023}. The gold standard for diagnosis remains manual microscopy of Giemsa-stained blood smears, which requires trained microscopists who are scarce in endemic regions~\cite{tangpukdee2009}. Rapid diagnostic tests (RDTs) offer an alternative but cannot quantify parasitemia or reliably distinguish \textit{Plasmodium} species, both critical for treatment decisions~\cite{murray2008}.

Deep learning has shown promising results for malaria detection~\cite{rajaraman2018,fuhad2020,manescu2020}, but most ML targeting malaria has been developed in an ML-centric, rather than clinic-centric perspective. For example, most systems require either cloud servers with reliable connectivity or high-performance compute hardware. In addition,
 the majority focus on single-cell classification rather than slide-level diagnosis, and few address parasitemia quantification with uncertainty estimates~\cite{poostchi2018}. Mobile-optimized detectors such as YOLO~\cite{ultralytics2024} combined with TensorFlow Lite~\cite{tflite2023} now enable on-device deployment, but deployment raises open challenges regarding  aggregating per-image detections into a reliable slide-level diagnosis, quantify uncertainty in the parasitemia estimate, and integrate human expert oversight efficiently.

Therefore we present a 
mobile system for malaria diagnosis, designed in consultation with a national health center (anonymized) with a large malaria-endemic catchment, to address several crucial clinical requirements. Contributions include:
\begin{enumerate}
    \item \textbf{Uncertainty-calibrated slide-level aggregation}
    via soft counting with validation-derived calibration, reducing systematic bias by a factor of 3 (vs hard counts)
    at all slide sizes while improving Pearson $r$ and RMSE across all configurations.
    \item \textbf{Quantitation confidence intervals (CIs)} via a computationally-efficient bootstrap resampling strategy over per-image statistics to compute 95\% CIs for slide-level parasitemia.
    \item \textbf{An early stopping
    criterion}, keyed to estimated parasitemia and uncertainty, signaling when sufficient images have been processed, which speeds clinical workflow.
    \item 
    \textbf{Human-in-the-loop (HITL)} interface with uncertainty-guided prioritization
    to support clinician oversight, review, and feedback (i.e. ML in a decision support role).
    \item \textbf{Lightweight ML models} to enable edge deployment in clinics without reliable internet or expensive compute resources.
\end{enumerate}

\section{Related Work}
\label{sec:related}

\paragraph{Deep Learning for Malaria Detection.}
Early CNN approaches classified individual cells~\cite{rajaraman2018,fuhad2020} without providing slide-level counts. Object detection frameworks (Faster R-CNN~\cite{ren2015}, YOLO~\cite{abdurahman2021,diker2022}) improved field-of-view analysis, but most work evaluates only per-image performance without slide-level aggregation for clinical diagnosis~\cite{yang2020,manescu2020}.
Current SOTA systems perform well, but require expensive scanning microscopes and compute power that are beyond the budgets of most front-line clinics \cite{horningWho55,das}.
The review by \cite{poostchi2018} describes further common problems with ML malaria efforts.

\paragraph{Mobile Deployment and Uncertainty.}
Quinn et al.~\cite{quinn2014} deployed thick-smear analysis on tablets but required server processing.
Newer generations of lightweight models, e.g.\ TensorFlow Lite~\cite{tflite2023}, enable on-device inference, yet no prior system (to our knowledge) combines multi-species detection with uncertainty-calibrated slide-level quantification on consumer hardware.

\section{Methods}
\label{sec:methods}

\subsection{System Architecture}
\label{sec:architecture}

\begin{figure}[t]
    \centering
    \includegraphics[width=\linewidth]{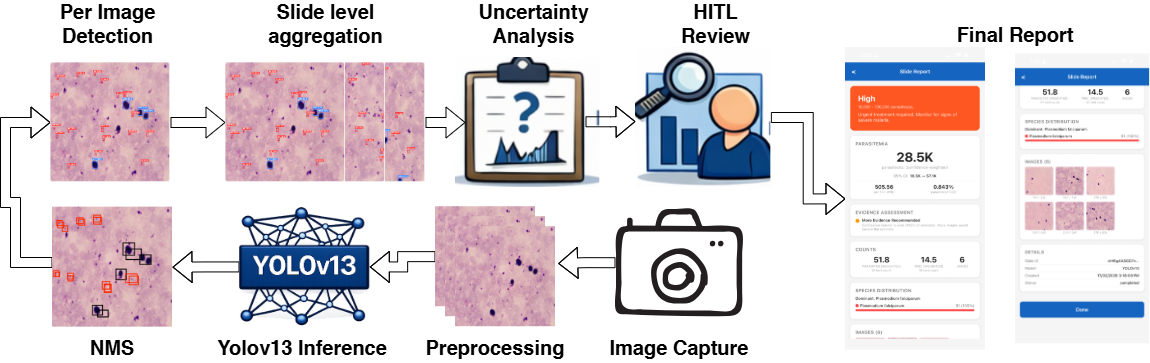}
    \caption{End-to-end pipeline from image capture to uncertainty-calibrated slide report.}
    \label{fig:pipeline}
\end{figure}

Our system is implemented as a cross-platform mobile application using React Native with custom native modules for TensorFlow Lite inference (Fig.~\ref{fig:pipeline}). The architecture comprises four stages. (1)~Image acquisition via camera or gallery import. (2)~On-device inference using YOLOv13n with native TFLite modules. (3)~Uncertainty-calibrated slide-level aggregation. (4)~Human-in-the-loop review with uncertainty-guided prioritization.
Images are stored locally; all processing occurs without network connectivity, with inference delegated to platform-specific TFLite interpreters (Kotlin on Android, Swift on iOS). Local image storage simplifies patient privacy logistics.

\subsection{Detection Model}
\label{sec:model}

We employ YOLOv13n~\cite{ultralytics2024} trained on Giemsa-stained thick blood smear images to detect five classes: \textit{P.~falciparum} (PF), \textit{P.~vivax} (PV), \textit{P.~malariae} (PM), \textit{P.~ovale} (PO), and white blood cells (WBC). The model is exported to TensorFlow Lite format at $2048 \times 2048$ input resolution. Input images undergo letterbox resizing with gray padding 
to preserve aspect ratio.
Post-processing implements DFL decoding (softmax over regression bins + dist2bbox), sigmoid on class logits, and per-class NMS (IoU 0.7, confidence threshold 0.25).

\subsection{Uncertainty-Calibrated Slide-Level Aggregation}
\label{sec:uncertainty}
To provide a patient-level recommendation, object-level detections must be aggregated across multiple images from the patient.

\subsubsection{Confidence-Weighted Soft Counting.}
\label{sec:soft_counting}
Traditional parasitemia estimation treats each detection as a binary count: $N_p = \sum_{i} \mathbb{1}[\text{class}_i \neq \text{WBC}]$. We instead weight each detection by its confidence:
\begin{equation}
    \tilde{N}_p = \sum_{i : \text{class}_i \neq \text{WBC}} c_i, \quad
    \tilde{N}_w = \sum_{i : \text{class}_i = \text{WBC}} c_i
    \label{eq:softcount}
\end{equation}
where $c_i \in [0, 1]$ is the confidence of detection $i$.
This approach naturally down-weights borderline detections while fully counting high-confidence ones. 
We note that WBC quantification functionality is vital because examined blood volumes (and thus parasitemias) are determined by WBC count \cite{whoMicroscopy,whoMicroscopyQuantSOP}:
\begin{equation}
    \tilde{P} = \frac{\tilde{N}_p}{\tilde{N}_w} \times 8{,}000 \quad \text{(parasites/\textmu L)}
    \label{eq:parasitemia}
\end{equation}
Since $c_i \in (0,1)$, raw soft counts systematically undercount relative to binary ground-truth annotations. We correct for this by deriving calibration factors on the validation set only: $\alpha_p = N_p^{\text{GT}} / \tilde{N}_p^{\text{val}} = 1.306$ for parasites and $\alpha_w = N_w^{\text{GT}} / \tilde{N}_w^{\text{val}} = 1.128$ for WBCs (at confidence threshold 0.05). All slide-level estimates use calibrated counts $\hat{N} = \alpha \cdot \tilde{N}$. Both $\alpha$ values are locked after validation and applied unchanged to the test set, ensuring no data leakage.

\subsubsection{Bootstrap Confidence Intervals.}
Uncertainty estimates are a valuable tool for clinical interpretation of algorithm results. To quantify quantitation reliability, we use bootstrap resampling~\cite{efron1993}: given $K$ analyzed images with per-image soft counts $(\tilde{n}_p^{(k)}, \tilde{n}_w^{(k)})$, for each of $B = 1{,}000$ iterations we draw $K$ images \emph{with replacement}, sum the soft counts, and compute parasitemia via Eq.~\ref{eq:parasitemia}. The 95\% CI is defined by the 2.5th and 97.5th percentiles of the resulting distribution. A narrow CI indicates stable estimation; a wide CI signals that more images are needed.

\subsubsection{Evidence Sufficiency Criterion.}
High patient loads require minimizing image collection time and time-to-result. To enable this on a per-patient basis, the system monitors bootstrap CI width during image acquisition and determines when enough images have been analyzed. Two conditions must hold: (1)~at least $K_{\min} = 3$ images analyzed, and (2)~CI precision:
\begin{equation}
    \tilde{P}_{\text{upper}} - \tilde{P}_{\text{lower}} < \max\!\left(0.3 \times \tilde{P},\; 500\right)
    \label{eq:evidence}
\end{equation}
The 30\% relative threshold ensures proportional precision for high-parasitemia cases, while the absolute floor of 500~parasites/\textmu L prevents the criterion from being unreachable in low-parasitemia cases with high Poisson variability \cite{delahuntPoisson}.

\subsubsection{Human-in-the-Loop Review.}
\label{sec:hitl} 
Clinician interaction and oversight are essential for accountability and quality control. To enable this, the HITL interface (Fig.~\ref{fig:pipeline}) supports soft-deletion of false positives, species re-labeling, and addition of missed detections via tap-to-place. These edits trigger live re-aggregation,  allowing the clinician to correct algorithm outputs.

\subsubsection{Uncertainty-Guided Prioritization.}
To focus clinician's review, each image receives a priority score to surface the most uncertain fields (i.e. those most likely to contain algorithm errors):
\begin{equation}
    s_k = 0.4\,(1 - \bar{c}_k) + 0.3\,\hat{d}_k + 0.2\,m_k + 0.1\,(1 - \mathbb{1}[n_w^{(k)} > 0])
    \label{eq:priority}
\end{equation}
where $\bar{c}_k$ is mean detection confidence, $\hat{d}_k = |n_p^{(k)} - \bar{n}_p|/(\bar{n}_p + \epsilon)$ is the normalized count deviation from the slide mean, $m_k$ indicates multi-species detections, and the final term flags absent WBC detections. Images are grouped into high ($s_k \geq 0.5$), medium ($\geq 0.25$), and low priority tiers.

\section{Experimental Setup}
\label{sec:experiments}

\paragraph{Dataset}
The dataset consists of 2,739 Giemsa-stained blood smear images from a national reference laboratory (anonymized),
spanning four \textit{Plasmodium} species and WBCs: PF (838 images, 7,568 instances), PM (834 images, 1,802 instances), PO (893 images, 2,353 instances), and PV (174 images, 669 instances). Images were auto-oriented, resized to $2048 \times 2048$ via letterbox padding, and split 70/15/15\% for train/val/test. To counter PV class imbalance (6.4\% of images), we applied targeted augmentation (rotation, hue/saturation/brightness variation) to the training set only.

\paragraph{Training Configuration}
YOLOv13n was trained on a single NVIDIA H100 GPU for 70 epochs (batch 6, SGD lr=0.01, CIoU + DFL loss, mosaic augmentation).

\paragraph{Evaluation Metrics}
\label{sec:eval_metrics}
We report per-class precision, recall, mAP@50, mAP@75, and mAP@50--95. Per-image count accuracy compares model-predicted counts against ground-truth (GT) annotations across all 410 test images via Pearson $r$, MAE, and RMSE. GT annotations are binary integer counts, so hard counts (threshold 0.25) enable direct comparison; soft counting is evaluated at the slide level where both methods produce a comparable parasitemia estimate.

Slide-level accuracy uses simulated slides: for $K \in \{5, 10, 15\}$ images/slide, we sample 50 slides (fixed seed, without replacement from the 410 test images), compute parasitemia via the WHO formula, and measure empirical 95\% CI coverage.

\paragraph{Mobile Performance}
On-device benchmarking was performed on a consumer smartphone (A15 Bionic, 6\,GB RAM) running iOS, CPU backend with 2 threads, across all 410 test images.

\section{Results}
\label{sec:results}

\subsection{Detection Performance}

Table~\ref{tab:detection} summarizes detection performance on the test set.
The model achieves an overall mAP@50 of 0.863 and mAP@50--95 of 0.626. PF yields the lowest mAP@50 (0.765) due to its small ring-form morphology, while PV achieves the highest parasite mAP@50 (0.903) despite being the most underrepresented class, validating the targeted augmentation strategy. WBC detection achieves the highest recall (0.936) and mAP@50--95 (0.760), which is critical due to WBCs' role in the WHO parasitemia formula.

\begin{table}[t]
    \centering
    \caption{YOLOv13n detection performance on the test set (410 images).}
    \label{tab:detection}
    {\small
    \begin{tabular}{lccccc}
        \toprule
        \textbf{Class} & \textbf{Precision~} & \textbf{~Recall~} & \textbf{~mAP@50~} & \textbf{~mAP@75~} & \textbf{~mAP@50--95} \\
        \midrule
        All classes      & 0.813 & 0.824 & 0.863 & 0.671 & 0.626 \\
        \midrule
        PF               & 0.749 & 0.665 & 0.765 & 0.577 & 0.508 \\
        PM               & 0.818 & 0.840 & 0.871 & 0.649 & 0.631 \\
        PO               & 0.842 & 0.816 & 0.880 & 0.783 & 0.705 \\
        PV               & 0.844 & 0.860 & 0.903 & 0.525 & 0.525 \\
        WBC              & 0.814 & 0.936 & 0.896 & 0.824 & 0.760 \\
        \bottomrule
    \end{tabular}}
\end{table}
\subsection{Slide-Level Parasitemia Estimation}
Table~\ref{tab:counting} presents per-image count accuracy. The overall parasite count correlation is $r = 0.812$ with a 5.4\% over-count (1,785 predicted vs.\ 1,694 GT), while WBC correlation is $r = 0.947$ with a 20.4\% over-count (615 vs.\ 511).

\begin{table}[t]
    \centering
    \caption{Per-image count accuracy: predictions vs.\ ground truth (410 test images).}
    \label{tab:counting}
    \begin{tabular}{lcccc}
        \toprule
        \textbf{Class} & \textbf{GT total~} & \textbf{~Pred total~} & \textbf{~Pearson $r$~} & \textbf{~MAE} \\
        \midrule
        PF   & 1,085 & 1,144 & 0.850 & 1.08 \\
        PM   & 200   & 215   & 0.934 & 0.12 \\
        PO   & 316   & 330   & 0.887 & 0.21 \\
        PV   & 93    & 96    & 0.948 & 0.06 \\
        WBC  & 511   & 615   & 0.947 & 0.33 \\
        \midrule
        All parasites & 1,694 & 1,785 & 0.812 & 1.43 \\
        \bottomrule
    \end{tabular}
\end{table}

Table~\ref{tab:parasitemia} compares hard counting and calibrated soft counting across slide sizes. Soft counting consistently outperforms the hard-count baseline: Pearson $r$ improves at every $K$ (0.889 vs.\ 0.845, 0.951 vs.\ 0.905, 0.874 vs.\ 0.804), RMSE is lower at every $K$, and systematic bias is reduced from 9.4--15.0\% (hard) to ${\leq}4.5$\% (soft). Hard counting over-counts WBCs by 20.4\% vs.\ 5.4\% for parasites, deflating the parasites/WBC ratio and causing underprediction of parasitemia; soft counting down-weights borderline WBC detections proportionally, correcting this imbalance. The non-monotonic hard-count Pearson $r$ across $K$ (0.845, 0.905, 0.804) reflects sampling variance in 50 simulated slides; RMSE and MAE decrease monotonically, confirming consistent accuracy gains with more images.

\begin{table}[t]
    \centering
    \caption{Slide-level parasitemia estimation: soft counting vs.\ hard-count baseline (50 simulated slides per configuration, seed=42). MPE = $(\hat{P}_{\text{GT}}{-}\hat{P})/\hat{P}_{\text{GT}}{\times}100$; positive = underprediction. Hard-count threshold (0.25) and soft-count calibration factors ($\alpha_p{=}1.306$, $\alpha_w{=}1.128$, threshold 0.05) were fixed on the validation set.}
    \label{tab:parasitemia}
    {\small
    \begin{tabular}{cccccc}
        \toprule
        \textbf{~K~} & \textbf{~Method~~} & \textbf{~Pearson $r$~~} & \textbf{~MPE(\%)~~} & \textbf{~RMSE~~} & \textbf{MAE} \\
        \midrule
        \multirow{2}{*}{5}
          & Hard          & 0.845 & $+9.4$  & 22,346 & 13,528 \\
          & \textbf{Soft} & \textbf{0.889} & $\mathbf{-0.5}$ & \textbf{18,477} & \textbf{11,073} \\
        \midrule
        \multirow{2}{*}{10}
          & Hard          & 0.905 & $+15.0$ & 15,032 &  8,667 \\
          & \textbf{Soft} & \textbf{0.951} & $\mathbf{+4.5}$  & \textbf{10,679} & \textbf{6,227} \\
        \midrule
        \multirow{2}{*}{15}
          & Hard          & 0.804 & $+9.7$  &  7,415 &  5,118 \\
          & \textbf{Soft} & \textbf{0.874} & $\mathbf{-1.8}$  & \textbf{5,391} & \textbf{3,661} \\
        \bottomrule
    \end{tabular}}
\end{table}

\subsection{Confidence Interval Calibration}
Table~\ref{tab:ci_calibration} reports bootstrap CI calibration. Coverage increases monotonically from 80\% (K=5) to 94\% (K=15), approaching the nominal 95\% level. The majority of misses at K=5 involve slides with ${\leq}14$ predicted WBCs, confirming WBC scarcity as the primary failure mode. Mean CI width narrows from 80,718 to 37,801~p/\textmu L with more images.

\begin{table}[t]
    \centering
    \caption{Bootstrap 95\% CI calibration (50 simulated slides, 1000 iterations each). CI width in p/\textmu L.}
    \label{tab:ci_calibration}
    \begin{tabular}{cccc}
        \toprule
        \textbf{Images/slide~~~} & \textbf{Coverage~~~} & \textbf{CI width (mean)~~~} & \textbf{CI width (median)} \\
        \midrule
        5  & 80.0\% (40/50) & 80,718 & 49,345 \\
        10 & 88.0\% (44/50) & 48,871 & 36,946 \\
        15 & 94.0\% (47/50) & 37,801 & 29,766 \\
        \bottomrule
    \end{tabular}
\end{table}

\subsection{Evidence Sufficiency and HITL Prioritization}
The evidence sufficiency module monitors CI width during acquisition and alerts the microscopist when additional images are needed; its clinical motivation is to terminate acquisition as soon as the estimate is reliable, reducing unnecessary image capture in resource-limited settings. In our batch simulation, CI widths remain wide relative to the stopping threshold at all tested slide sizes, indicating that the criterion is designed for scenarios where parasitemia is high enough to yield narrow per-image counts quickly — prospective evaluation with real-time sequential acquisition is needed to quantify fields-of-view saved. The HITL module surfaces low-confidence, outlier-count, and WBC-absent images for priority review. The HITL interface was designed in direct response to clinical requirements for expert oversight, enabling corrections to false positives, species re-labeling, and missed detections with live re-aggregation. Quantitative user-study evaluations of both modules are important future work. 

\subsection{Mobile Performance}
Table~\ref{tab:performance} summarizes on-device performance. The per-image pipeline averages $10.27 \pm 1.65$~s, dominated by model invocation ($7.90 \pm 1.29$~s). Post-processing (DFL decode + NMS) takes $220 \pm 32$~ms; bootstrap CI is negligible ($<50$~ms).

\begin{table}[t]
    \centering
    \caption{On-device performance (A15 Bionic SoC, CPU, 2 threads, $n = 410$).}
    \label{tab:performance}
    \begin{tabular}{lc||lc}
        \toprule
        \textbf{Metric} & \textbf{Value} & \textbf{~Metric} & \textbf{Value} \\
        \midrule
        Model input size           & $2048 \times 2048~~~$ & ~Preprocessing time         & $2.15 \pm 0.35$ s \\ 
        Peak memory usage          & 1.7 GB & ~Model invoke time & $7.90 \pm 1.29$ s \\
        Mean detections/image      & 5.9 & ~Post-processing time & $220 \pm 32$ ms \\
        Model size (TFLite)        & 48 MB & ~Total per-image pipeline~   & $10.27 \pm 1.65$ s \\
        \bottomrule
    \end{tabular}
\end{table}

\section{Discussion}
\label{sec:discussion}

We presented an on-device mobile system for multi-species malaria detection
designed to specifically address key use case constraints described by our healthcare collaborators.
The system achieves $r = 0.812$ per-image and $r = 0.951$ slide-level parasite count correlation, operating entirely offline on consumer smartphones for deployment in resource-limited settings.

\paragraph{Limitations.}
The $2048 \times 2048$ input results in 1.7\,GB peak memory and ${\sim}10$\,s pipeline time, limiting deployment to higher-end devices. 
The CPU-only backend was necessitated by hardware delegate incompatibilities (CoreML crashes, Metal hangs); resolving these would substantially reduce inference time. 
Simulated slides sample images without patient grouping, as patient identifiers were unavailable under the clinical partners' data governance policy, making slide-level correlations a lower bound on performance with true per-patient slides. 
Our evaluation is limited to thick smears from a single reference laboratory; extension to thin smears and cross-site evaluation remains future work, along with prospective clinical validation against expert microscopy across diverse settings. \\




\noindent\textbf{Prospects of Application.}
Clinics treating malaria in low-resource settings have operational requirements which are currently under-studied in the ML literature, but which carry strong implications for ML development teams. Guided by our healthcare system collaborators, this paper highlights these under-represented but crucial design constraints that gate deployment, and offers ML methods to address them. 


%
%
\bibliographystyle{splncs04}
\bibliography{references}

\end{document}